\documentclass{bmvc2k}

\usepackage{comment}
\usepackage{multirow}
\usepackage{tabularx}
\usepackage{makecell}
\usepackage{amsmath}
\usepackage{amssymb}
\usepackage{stmaryrd}
\usepackage{amsfonts}
\usepackage{pifont} 
\usepackage{stmaryrd}

\newcommand{\cmark}{\ding{51}}
\newcommand{\xmark}{\ding{55}}

\DeclareMathOperator{\card}{Card}

\title{Learning Finer-class Networks for Universal Representations}

\addauthor{Julien Girard}{julien.girard2@cea.fr}{12}
\addauthor{Youssef Tamaazousti}{youssef.tamaazousti@cea.fr}{123}
\addauthor{Herv\'e Le Borgne}{herve.le-borgne@cea.fr}{2}
\addauthor{C\'eline Hudelot}{celine.hudelot@centralesupelec.fr}{3}

\addinstitution{
 Both authors contributed equally.
}
\addinstitution{
 CEA LIST\\
 Vision Laboratory,\\ 
 Gif-sur-Yvette, France.
}
\addinstitution{
 CentraleSup\'elec,\\
 MICS Laboratory,\\ 
 Ch\^{a}tenay-Malabry, France.
}

\runninghead{Girard et al.}{Finer-Class Networks}

\begin{document}

\maketitle

\begin{abstract}
Many real-world visual recognition use-cases can not directly benefit from state-of-the-art CNN-based approaches because of the lack of many annotated data. 
The usual approach to deal with this is to transfer a representation pre-learned on a large annotated source-task onto a target-task of interest. 
This raises the question of how well the original representation is ``universal'', that is to say directly adapted to many different target-tasks. 
To improve such universality, the state-of-the-art consists in training networks on a diversified source problem, that is modified either by adding generic or specific categories to the initial set of categories. 
In this vein, we proposed a method that exploits finer-classes than the most specific ones existing, for which no annotation is available. We rely on unsupervised learning and a bottom-up split and merge strategy. 
We show that our method learns more universal representations than state-of-the-art, leading to significantly better results on 10 target-tasks from multiple domains, using several network architectures, either alone or combined with networks learned at a coarser semantic level. 
\vspace{-0.5cm}
\end{abstract}

%==============================
% INTRODUCTION
%==============================
\section{Introduction}
\label{sec:intro}
\vspace{-0.1cm}

The state-of-art performances in visual recognition obtained by Convolutional Neural Networks (CNNs) are subject to the availability of a large set of annotated training data to learn the model. Since it is rarely the case for many practical tasks of interest (\textit{target-tasks}), one usually adopts a transfer-learning approach~\cite{oquab2014learning,razavianASC14,popescu15arxiv} which relies on a CNN pre-trained on a \textit{source task} with sufficient annotated data (often ImageNet~\cite{russakovsky2014imagenet}) then further truncated to provide the representations of the samples of target-task. 
Then, even with few annotated data, this last can usually be learned with a linear classifier. 
Such approaches raise the question of the similarity of the \textit{source-task} on which the representation has been learned and the target-task on which it is used. 
Although this similarity is not easy to formalize, one has the intuition that the closer the both tasks the better the representation will be adapted to the target-task. %(and thus the performances on it). 
This consideration leads to several methods that tend to obtain more \textit{universal} representations~\cite{bilen2017universal,conneau2017supervised,kokkinos2017ubernet,rebuffi2017learning,tamaazousti2017mucale_net,tamaazousti2018universal}, that is to say that are more adapted to a large set of diverse target-tasks, in a transfer-learning scenario. 

The general idea of these methods is to diversify the classification problem of the source-task in order to obtain more features, able to adequately represent new target-datasets, from more domains, in a larger context. 
All these approaches vary the problem by creating new categories having an existing label. 
However, most of them studied the effect of adding categories extracted from ImageNet, either \textit{generic} categories~\cite{krizhevsky2012imagenet,mettesicmr16,tamaazousti2017mucale_net, tamaazousti2018universal} or \textit{specific} ones~\cite{zhou2014learning,azizpour2015generic,bilen2017universal,rebuffi2017learning}, that are at the bottom of a hierarchy such as ImageNet, except~\cite{joulin2015learning,vo15cbmi,vo2017harnessing} that use web annotations with noisy labels. 
In general, the usage of specific categories tends to provide better performances than generic ones~\cite{bilen2017universal,rebuffi2017learning}, although combining them can significantly boost the universalizing capacity of the CNN~\cite{tamaazousti2017mucale_net,tamaazousti2018universal}. 
Yet, even for the most specific categories, it is plausible that it exists a variety of semantics within the class that is not explored (\textit{e.g.}, one could imagine to split the object-class according to the different poses of the object). 
Clearly, the limiting point is the availability of such finest annotation (\textit{e.g.}, poses, contexts, attributes) for existing specific classes.
  
In this article, we argue that exploring \textit{finer} classes than the most \textit{specific} existing ones, can significantly increase the diversity of the problem, therefore improve the universality of the representation learned in the internal layers of the CNN. 
The main difficulty is the lack of annotation below the most specific levels. 
We propose to rely on unsupervised learning (clustering) to determine these finer categories within each specific category. 
Our contribution is three-fold. 
First, we show that the use of finer categories rather than the most specific ones to learn CNNs, improves the universality of the resulting representation, even when the finer classes are determined \textit{randomly} within each specific class. 
Second, the usage of a K-means based approach leads to slightly better results although the resulting clusters are strongly imbalanced. To fix this, our core contribution splits and merges the specific categories to automatically determine better balanced finer classes, leading to better results. 
Last, we show that CNNs learned with our approach provide a better complementary to standard CNN representations than those learned on generic categories. 

Let note that if the target-task has enough data, the representation can be adapted to the target-task by fine-tuning. This is nevertheless out of the scope of this work, because it is a \textit{complementary} process to the transfer-learning in itself, that will always improve the performances, and especially, because fine-tuning modifies the representations, which leads to a bias that \textit{hides the real ability} of a universalizing method~\cite{huh2016makes}. 
Hence, in this paper, we are only interested into studying the universality of the representations, independently of many possible refinements of a full adaptation method on each target-task. 

Previous works~\cite{dong2013subcategory,dong2015looking,xiang2017subcategory,chen2017s} exploited sub-categories in the context of visual recognition. 
In~\cite{dong2013subcategory,dong2015looking}, an object instance affinity graph is computed from intra-class similarities and inter-class ambiguities then the visual subcategories are detected by the graph shift algorithm. 
The process is nevertheless quite computationally demanding and applied to object detection on small target datasets only. 
In~\cite{xiang2017subcategory,chen2017s}, subcategories that are learned from extrapolated feature maps and fine-tuned on a target-dataset, are used within a CNN to improve region proposal for object detection. 
To the best of our knowledge, our paper is the first to propose the usage of subcategories determined by unsupervised learning on a source-task, in order to improve universality of representations. 
More related to universality, \cite{bilen2017universal,rebuffi2017learning,rebuffi2018efficient} added annotated data from more domains as well as domain-specific neurons to an initial set of domain-agnostic ones.  
Contrary to them, our method only modifies the source-problem at zero cost of annotation. 
Our work is closer to the approach of~\cite{tamaazousti2017mucale_net,tamaazousti2018universal} that proposes to relabel specific categories into generic ones (that match the upper categorical-levels), to learn an additive CNN with the same architecture. 
Nevertheless, our approach is interested into the ``opposite way'', that is, creating finer classes than those at the bottom of a hierarchy (ImageNet), for which no annotation exists and thus their method can not be applied.

We evaluated our proposal on the problem of universality, that is, in a transfer-learning scheme using multiple target-tasks (\textit{i.e.}, ten classification benchmarks from multiple domains, including actions, food, scenes, birds, aircrafts, etc.). 
In particular, in comparable settings (using ILSVRC as source-task and two architectures, AlexNet~\cite{krizhevsky2012imagenet} and DarkNet~\cite{redmon2017yolo9000}), we showed that our method outperforms state-of-the-art ones.  

%==============================
% PROPOSED METHOD
%==============================
\section{Proposed Method}
\vspace{-0.2cm}

We propose a new universalizing method that consists in training a network on a set of categories that are \textit{finer} than those of the \textit{finest}-level of a hierarchy (\textit{e.g.}, ImageNet hierarchy or any set of categories). 
In Sec.~\ref{sec:finet_splitting_random_clustering}, we start by describing its general principle as well as two baselines that splits them either randomly or by clustering their features. 
With such baseline, the number of finer classes must be a priori fixed, thus we propose a ``bottom-up clustering-based merging''  approach that determines a better splitting automatically (Sec.~\ref{sec:bucbam}). 
Furthermore, we propose to combine the features learned on the specific categories and those learned on the finer ones to get an even more universal representation (Sec.~\ref{sec:spefinet}). 

\begin{figure*}[tb!]
\begin{center}
	\includegraphics[width=12.0cm]{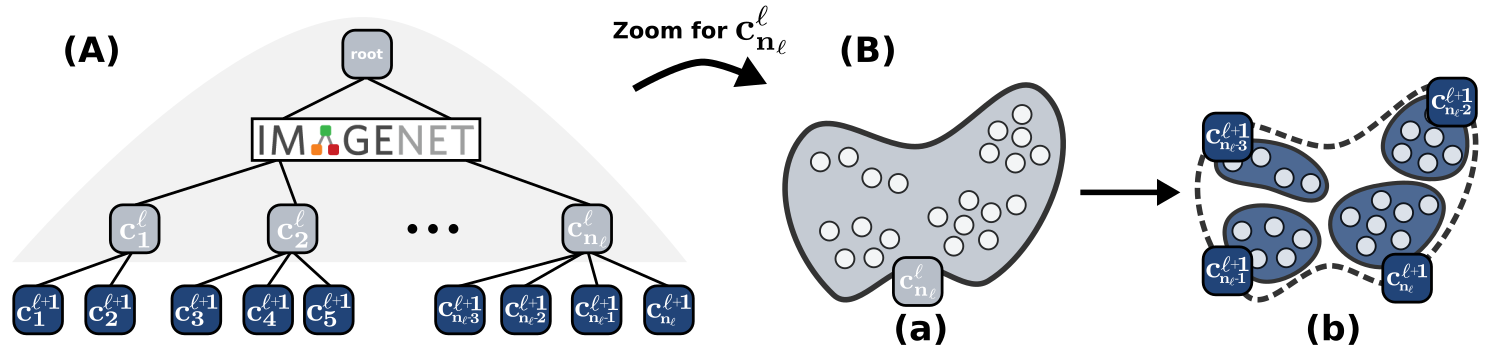}
\end{center}
\vspace{-0.5cm}
\caption{
Illustration of our splitting principle that determines a \textit{finer}-level $\ell$+1 containing $n_{\ell +1}$ \textit{finer}-classes (blue nodes in \textbf{A}) from the \textit{finest}-level $\ell$ of a hierarchy (here, ImageNet), containing $n_{\ell}$ \textit{specific} categories that the leaf gray nodes in \textbf{(A)}. 
Each finer-class $c_i^{\ell +1}$ ($i \in [n_{\ell +1}]$) is related to a specific-class $c_j^{\ell}$ ($j \in [n_{\ell}]$) through ``is-a'' relations, since the $c_i^{\ell +1}$ classes are obtained from \textit{each} $c_j^{\ell}$ class. 
In \textbf{(B)}, we focus on the particular specific class $c_{n_l}^{\ell}$ (\protect \includegraphics[width=2.0ex,height=1.5ex]{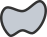}) and its image-representations (\protect \includegraphics[width=1.5ex,height=1.5ex]{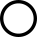}), to determine the finer-classes $\{c_{n_l-3}^{\ell +1}, c_{n_l-2}^{\ell +1}, c_{n_l-1}^{\ell +1}, c_{n_l}^{\ell +1}\}$. 
In (b), it is splitted into $K$ (here $K$$=$$4$) groups (\protect \includegraphics[width=2.0ex,height=1.5ex]{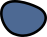}) corresponding to the finer-classes $c_j^{\ell +1}$. 
}
\vspace{-0.3cm}
\label{fig:finer-level_finer-classes}
\end{figure*}

%--------------------------
% FINER CLASSES BY RANDOM/CLUSTER-SPLITTING
%--------------------------
\subsection{FiNet: Network Trained on Finer-Classes}
\label{sec:finet_splitting_random_clustering}
\vspace{-0.2cm}

The leaf nodes of the ImageNet hierarchy represents the \textit{finest} or most  \textit{specific} categories that are annotated. 
More generally, this is the case for the set of categories of any classification dataset. 
To go towards our goal of automatically obtaining finer categories (without annotations) from the finest ones, a baseline approach consists in using a random partitioning of the specific categories or a simple clustering-based approach of their image-features. 
The first baseline, randomly assigns every image of a specific category to one of $K_i$ clusters. 
The second one, first, learns a CNN (noted \textbf{SpeNet}) on the specific categories, uses one of its layer as features-extractor for every image, then determines $K_i$ clusters using K-means on these vectors. 
A more sophisticated way is our final method that is presented in Sec.~\ref{sec:bucbam}. 
Note that, in all cases, the splitting is performed on specific categories, that already contains quite similar samples/vectors. 
Once the finer classes obtained, we train another network (denoted \textbf{FiNet}) on the \textit{same} images used to train SpeNet, but labeled among the obtained \textit{finer}-classes. The whole set of finer-classes forms the new finest-level of the hierarchy. 
Our general principle is illustrated in Fig.~\ref{fig:finer-level_finer-classes} and presented more formally below. 

Let us consider a semantic hierarchy with hyponymy relations, that is to say a set of categories organized according to ``is-a'' relations (\textit{e.g.}, ImageNet~\cite{deng09imagenet} hierarchy). 
This hierarchy denoted $\mathcal{H}_{\ell}=(\mathcal{C}_{\ell},E)$ is a directed acyclic graph of $\ell$ levels of nodes, with $\mathcal{C}_{\ell}$ being all the nodes and $E$ the set of directed edges between the nodes. 
Each node $c_i^{\ell}\in \mathcal{C}_{\ell}$ corresponds to the $i$-th category at level $\ell$ in $\mathcal{H}_{\ell}$ and $n_{\ell}$ is the number of categories at level $\ell$. 
A hierarchy-edge $(c_i^{\ell},c_j^{\ell +1})\in E$ indicates that class $c_i^{\ell}$ subsumes class $c_j^{\ell +1}$. 
Let us also consider an initial dataset $\mathcal{D}_{N}^{\ell}$ containing a set of $N$ images labeled among the categories at level $\ell $ and let us denote $N_{i}^{\ell}$ the number of images labeled among the $i$-th category at level $\ell$. 
Note that, $N = \sum_{i=1}^{n_l} N_i^{\ell}$. 
Each image $\textbf{I}_i^j\in \mathcal{D}_N^{\ell}$ of the dataset, is associated to a given category $c_j^{\ell}$ for $j\in [n_{\ell}]$\footnote{Let $[n]$ denotes $\llbracket1 ,n \rrbracket$, in all the paper.}. 
Let us denote $\textbf{X}_i^{\ell, L}\in \mathbb{R}^{d_{\ell}}$ the representation of an image $\textbf{I}_i$ extracted from layer $L$ of the network trained on $\mathcal{D}_N^{\ell}$ (\textit{i.e.}, SpeNet). 
Let also $\mathcal{X}_i^{\ell} = \{ \textbf{X}_j^{\ell,L}\}_{j=1}^{N_i^{\ell}}$ being the set of features extracted from all the images belonging to the category $c_i^{\ell}$. 

In order to construct the $K_i$ finer-categories $\{c_j^{\ell +1}\}_{j=1}^{ n^{\ell+1}}$ of \textit{each} category of the previous level $c_i^{\ell}$, we apply a clustering algorithm (\textit{e.g.}, K-means, MeanShift or BUCBAM presented in Sec.~\ref{sec:bucbam}) on $\mathcal{X}_i^{\ell}$ (that builds $K_i$ centroids) where the feature vector $\textbf{X}_j^{\ell,L}$ of each image $\textbf{I}_j$ ($j\in [N_i^{\ell}]$) is assigned to the nearest centroid (hard-coding~\cite{liu2011defense}), which forms the $K_i$ finer classes. 
This process is applied for all $i\in [n_{\ell}]$ which gives the $n_{\ell +1}=K_i\times n_{\ell}$ finer classes, that forms the nodes of the finer level $\ell +1$. 
This latter results in a new dataset $\mathcal{D}_N^{\ell +1}$, for which each image $\textbf{I}_i^j\in \mathcal{D}_N^{\ell +1}$ is associated to a given category $c_j^{\ell +1}$ for $j\in [n_{\ell +1}]$. 
Note that by construction, every $c_i^{\ell}$ subsumes all its finer-categories $\{c_{j}^{\ell +1}\}_{j=1}^{K_i}$, thus we have ($c_i^{\ell}, c_{j}^{\ell +1})_{(i,j)\in [n_{\ell}]\times[K_i]}$. 
The whole process results in a new hierarchical level $\ell +1$ that forms the new hierarchy $\mathcal{H}_{\ell+1}=(\mathcal{C}_{\ell+1},E')$ with $\mathcal{C}_{\ell+1} = \mathcal{C}_{\ell}\cup \{ c_{j, j\in[K_i]}^{\ell +1} \}_{i\in [n_{\ell}]}$.  
$E'$ corresponds to the union of $E$ and the edges that connect each category $c_i^{\ell}$ to its $K_i$ finer ones $c_j^{\ell +1}$. 
It is important to point out that, depending on the clustering algorithm, $K_i$ will depend on $c_i^{\ell}$ or be the \textit{same} for all categories. This is discussed in the next section. 

The new dataset $\mathcal{D}_{N}^{\ell+1}$ is used to train (softmax cross-entropy loss minimized by SGD) the FiNet network, which has $n_{\ell +1}$ neurons on its last layer. 
FiNet is then used as features-extractors for the images $\textbf{I}_i$ of the target-tasks: $\textbf{X}_i^{\ell +1, L}=\Phi^{\ell +1}_L(\textbf{I}_i)$. 

\begin{figure*}[tb!]
\begin{center}
	\includegraphics[width=9.0cm]{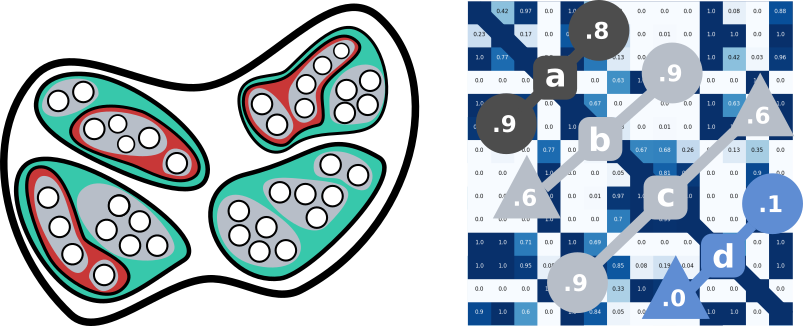}
\end{center}
\vspace{-0.5cm}
\caption{
On the left, given image-representations (\protect \includegraphics[width=1.5ex,height=1.5ex]{legends/circle.png}) labeled among a specific class (\protect \includegraphics[width=2.0ex,height=1.5ex]{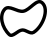}), 
BUCBAM first performs a clustering with many clusters (\protect \includegraphics[width=2.0ex,height=1.5ex]{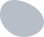}), then attaches small clusters to bigger closest ones (\protect \includegraphics[width=2.0ex,height=1.5ex]{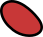}) and lastly, merges similar ones (\protect \includegraphics[width=2.0ex,height=1.5ex]{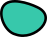}), w.r.t certain strategies. 
On the right, we describe the latter merging strategies. 
Given the similarity matrix $\textbf{M}_j$ (high values: blue, low: white) for one specific class $c_j^{\ell}$, 
BUCBAM-SS merges \textit{only} clusters that are reciprocally highly similar to each other,  respecting constraint (a) with high scoring values in $(\textbf{M}_j)_{k,l}$ and $(\textbf{M}_j)_{l,k}$). 
BUCBAM-AS \textit{also} merges asymmetrically similar clusters (those that respect constraints (a), (b) and (c)). 
In both cases, dissimilar clusters (d) are let disjoint. 
}
\vspace{-0.5cm}
\label{fig:bottom_up_clustering_based_merging}
\end{figure*}

%--------------------------
% BOTTOM-UP CLUSTERING-BASED MERGING
%--------------------------
\subsection{Bottom-Up Clustering-Based Merging}
\label{sec:bucbam}
\vspace{-0.2cm}
We empirically observed (see Fig.~\ref{fig:results}) that clustering approaches with a fixed $K_i$ for each category (\textit{e.g.}, Kmeans) usually leads to a FiNet that gives better universality results than approaches that adapt $K_i$ to each category (\textit{e.g.}, Affinity-Propagation). 
Indeed, this latter tends to provide a set of finer classes with many clusters containing  few images and a couple of clusters containing a large number of images, leading to an undesirable imbalanced dataset that penalizes the network training.  
Even if the use of fixed-$K_i$ clustering methods leads to more balanced data, it remains sub-optimal since it sets the \textit{same} amount of clusters for all specific categories ($\forall i \in [n_{\ell}], K_i = K$), while this may depend on the content of each category. 
Furthermore, in fixed-$K_i$ clustering methods, the $K$ value is an hyper-parameter that is cross-validated on the target-tasks, which are not accessible during the learning on the source-task, in the context of universality~\cite{tamaazousti2018universal}. 
Hence, the cross-validation of $K$ should be performed \textit{only} with the \textit{source}-task, which is not trivial (optimal $K$ on the source-task not necessary  optimal one for the target-tasks). 
To overcome the drawbacks of fixed and adapted-$K_i$ clustering methods, we proposed an hybrid one called ``Bottom-Up Clustering-BAsed Merging'' (\textbf{BUCBAM}). 
It roughly starts with the clusters obtained by a fixed-$K_i$ clustering and \textit{automatically} sets the amount of clusters for \textit{each} specific category by enforcing a more balanced resulting set of finer-classes. 
Specifically, it consists in three main steps (illustrated on the left of Fig.~\ref{fig:bottom_up_clustering_based_merging}): (i) splitting the specific categories into $K$ clusters (with a large $K$); (ii) attaching small clusters to the closest bigger ones (to avoid imbalanced data); and (iii) merging the most similar ones, with respect to a proposed similarity criteria.  

Formally, BUCBAM starts with a \textit{large} amount of $K$ finer-classes per category $c_{j}$ with $j \in [n_{\ell}]^{\ell}$ and $K$ being the \textit{same} for all $c_j^{\ell}$. 
Let us assume we have $K(\in \mathbb{N}^*)$ finer-categories $\{c_i^{\ell +1}\}_{i\in [K_j]}$ obtained from the category $c_j^{\ell}$ of the previous level through a fixed-$Kçi$ clustering method.  
Let denote $\mathcal{X}_i^{\ell +1} = \{ \textbf{X}_i^{\ell,L}\}_{i\in [N_i^{\ell +1}]}$ the whole set of features extracted from the images of a given category $c_i^{\ell +1}$ through the SpeNet $\Phi^{\ell}_L$. 
Note that, $N_i^{\ell +1}$ corresponds to the amount of images in each $c_i^{\ell +1}$\footnote{For simplicity, we omit the power indices $\ell$, $\ell +1$ and $L$ in the following.}.  
The goal of BUCBAM is to get an amount of clusters $K_j$ depending on the images of \textit{each} category $c_j^{\ell}$. 
To do so, it first prunes out the \textit{small} clusters (\textit{i.e.}, all the $c_i$ such that $\forall i\in [K_j]$, $\card(c_i) < S\in \mathbb{N}^*$, with $S \ll N_i$), by re-assigning their samples $\textbf{I}_{k, k\in[N_i]}$ (that were assigned to $c_i$) to the category of the closest feature vector $\textbf{X}_{l, l \in [N_i]}^i = \mathcal{N}(\textbf{X}_{k, k\in [N_m]}^m)$, with $m \neq i$ and $\mathcal{N}(\cdot)$ being a function that provides the closest vector (\textit{e.g.}, k-NN algorithm with Euclidean distance) in the set of features $\{\mathcal{X}_j\}$ belonging to the \textit{other} and \textit{large} clusters (\textit{i.e.}, all $c_m$ with $\card(c_m) \geqslant S$).  
Pruning small clusters for all categories $c_j^{\ell}$, results in a set of $K_j^{\mathcal{P}}$ finer-classes $\{c_i\}_{i\in K_j^{\mathcal{P}}}$ per class $c_j^{\ell}$. 
The last step of BUCBAM is to merge the similar clusters. 
To do that, a classifier $\Psi_i$ is trained for each cluster $c_i$ -- using features of $c_i$ samples as positives and same amount of samples from a \textit{diverse} class $c_{d}$ as negatives -- and evaluated on the images of all other clusters. 
The diverse category $c_{d}$ is created by randomly picking elements equiprobably from all the categories $\{c_i^{\ell}\}_{i \in [n_{\ell}]}$. 
The evaluation of the classifiers provides a similarity matrix $\textbf{M}_j \in [0,1]^{K_i^{\mathcal{P}} \times K_i^{\mathcal{P}} }$ for each category $c_i^{\ell}$. 
This last is used to merge similar clusters and let dissimilar ones disjoint. 
More precisely, a first strategy is to consider clusters $c_i$ and $c_m$ \textit{symmetrically similar} (BUCBAM-SS) if: $\Psi_i(\textbf{X}_{k, k\in [N_m]}^m) > S_H$ and $\Psi_m(\textbf{X}_{k, k\in [N_i]}^i) > S_H$, with $m \neq i$ and $S_H \in [0,1]$ a \textit{high} score (close to 1). 
Another strategy is to consider, clusters \textit{asymmetrically similar} (BUCBAM-AS) if only one constraint is respected and the other is greater than $S_M$, with $S_M = S_H/2$ a \textit{medium} score. 
In both cases (that are illustrated in Fig.~\ref{fig:bottom_up_clustering_based_merging}), dissimilar clusters are desirably let disjoint. 
Merging similar clusters for all classes $c_j^{\ell}$, results in a set of $K_j^{\mathcal{M}}$ finer-classes $\{c_i\}_{i\in K_j^{\mathcal{M}}}$ per category $c_j$.
\vspace{-0.2cm}

%--------------------------
% BOTTOM-UP CLUSTERING-BASED MERGING
%--------------------------
\subsection{SpeFiNet: Combining Specific and Finer Features}
\label{sec:spefinet}
\vspace{-0.2cm}

Following the approach of~\cite{tamaazousti2018universal} -- which roughly consists in training initial features on an initial set of categories, then learning new features on new set of categories and finally combining initial and new features --, we propose to learn the new features with our FiNet (rather than a network trained on \textit{generic} categories~\cite{tamaazousti2017mucale_net,tamaazousti2018universal}) and combine them with the features of the initial SpeNet to get a representation even more universal. 
This method is denoted \textbf{SpeFiNet} in the following. 
Formally, the final SpeFiNet representation combines specific and finer features and is computed for an image $I_i$ of a target-task as: $ \textbf{X}_{i} = \mathcal{F}\big( \{\mathcal{Z}( \textbf{X}^{\ell}_{i}), \mathcal{Z}( \textbf{X}^{\ell +1}_{i})\} \big)$, where $\mathcal{F}$ is a fusion operator, and $\mathcal{Z}$ is a normalization function. 
In practice for the normalization and fusion, we respectively choose the L-infinite norm ($L$-$\infty$) and the concatenation. 
To the best of our knowledge, we are the first to propose to combine a SpeNet (trained on \textit{specific} categories) and a FiNet (trained on \textit{finer} categories) to get more universal representations. 
\vspace{-0.2cm}

%==============================
% EXPERIMENTAL RESULTS
%==============================
\section{Experimental Results}
\label{sec:settings}
\vspace{-0.2cm}

% UNIVERSALITY
\textbf{Universality}\\
Universalizing methods are evaluated in a transfer-learning scheme on multiple target-tasks~\cite{cer2018universal,conneau2017supervised,tamaazousti2018universal}. 
More precisely, a source-task is used to train a network that acts as a representation extractor on the data of the target-tasks. 
Each target-task is trained with a simple predictor on top of the representations extracted from the samples of the target-task. 
Note that, fine-tuning the representations on the target-tasks could always improve performances but induces a bias avoiding correct evaluation of universality~\cite{conneau2018senteval,huh2016makes,subramanian2018learning,tamaazousti2018universal}. 
Hence, following the literature, simple predictors that do not modify the representations learned on the source-task are used. 
In particular, here for the target-tasks, we used a classification task with datasets from multiple visual domains (presented below) and for the predictor, we used a one-versus-all SVM classifier for each class. 
Even if~\cite{conneau2018senteval,rebuffi2017learning,tamaazousti2018universal} initiated a work around universality evaluation, it seems to remain an open problem. 
Hence here, since we only have benchmarks that are evaluated in terms of accuracy and precision, we evaluate universalizing methods in terms of average of their performances on the multiple benchmarks.\\ 

% DATASETS 
\noindent\textbf{Datasets}\\ 
For the source-task, we used ILSVRC~\cite{russakovsky2014imagenet} and ILSVRC* (half of the former, detailed in~\cite{tamaazousti2017mucale_net}). 
For the target-tasks, we used ten datasets from multiple domains, including general objects (VOC07~\cite{everingham2010pascal}, NWO~\cite{nus-wide-civr09}, CA101~\cite{fei2006one}, CA256~\cite{griffin2007caltech}), scenes (MIT67~\cite{quattoni2009recognizing}), actions (stACT~\cite{yao2011human}), birds (CUB~\cite{wahCUB_200_2011}), plants (FLO~\cite{nilsback2008automated}), food (FOOD~\cite{bossard14}) and airplanes (AIRC~\cite{maji13fine-grained}). 
The characteristics of all the datasets are detailed in supplementary material.\\ 

% IMPLEMENTATION DETAILS
\noindent\textbf{Implementation Details}\\
Our method consists in the combination of a SpeNet and FiNet. 
For both networks, we used two architectures, namely the classical AlexNet~\cite{krizhevsky2012imagenet} and the deeper DarkNet~\cite{redmon2017yolo9000}. 
They are respectively trained on the images of ILSVRC* and ILSVRC. 
SpeNet is thus respectively trained to recognize $C=483$ and $C=1,000$ \textit{specific} categories. 
In contrast, FiNet is trained to recognize a set of $K_i\times C$ \textit{finer}-classes ($K_i$ depends on the splitting method), for which we used four variants: (i) random splitting with $K_i\in\{2,4,8,16\}$ fixed, denoted \textbf{Random-K}  (ii) K-means clustering with $K_i\in\{2,4,8,16\}$ fixed, denoted \textbf{Cluster-K} (iii) BUCBAM splitting with \textit{asymmetrically similar} clusters merging, denoted \textbf{BUCBAM-AS} and (iv) BUCBAM splitting with \textit{symmetrically similar} clusters merging, denoted \textbf{BUCBAM-SS}. 
Note that, the BUCBAM methods leads to a $K_i$ depending on the content of each category.  
In Sec.~\ref{sec:analysis}, we provides some statistics of the resulting dataset of each method, including the total amount of finer-classes. 
In Cluster-K and BUCBAM methods, we extract features from the penultimate layer to represent the samples of each class, which results in features of 4096 dimensions for AlexNet and 1000 for DarkNet. 
%The clustering is always performed with the k-means algorithm. 
Specific to BUCBAM, the $K$, $S$ and $S_H$ parameters are respectively set to 32, 15 and 0.8. Indeed, $K$ has to be large, and we found that as long as $K$ is larger than 20 our method provides the same splitting result. $S=15$ ensures to train a network with at least 15 images per class. We obtained similar results with $S=50$. The parameters $S_H$ is not critical since similar clusters generally provides very high (close to 1.0) classification scores. 
\vspace{-0.2cm}

\begin{table*}[tb!]
\centering
\bgroup
\def\arraystretch{1.2}
\resizebox{\textwidth}{!}{
\begin{tabular}{|l c c c c c c c c || c |}
\hline
\multirow{2}{*}{\textbf{Method}} & \textbf{VOC07} & \textbf{CA101} & \textbf{CA256} & \textbf{NWO} & \textbf{MIT67} & \textbf{stACT} & \textbf{CUB} & \textbf{FLO} & \multirow{2}{*}{\textbf{Avg}} \\
 & mAP & Acc. & Acc. & mAP & Acc. & Acc. & Acc. & Acc. & \\
\hline
\hline
\textbf{SpeNet (REFERENCE)} & 66.8  & 71.1  & 53.2  & 52.5 & 36.0 & 44.3  & 36.1 & 50.5 & 51.3 \\
\textbf{SPV$_{\mathbf{A}}^{\mathbf{spe}}$}~\cite{azizpour2015generic} & 66.6 & 74.7 & 54.7 & 53.2 & 37.4 & 45.1 & 36.0 & 51.9 & 52.4 \\
\textbf{SPV$_{\mathbf{G}}^{\mathbf{gen}}$}~\cite{mettesicmr16,tamaazousti2017vision} & 67.7 & 73.0 & 54.3 & 50.5 & 37.1 & 44.9 & 36.8 & 50.3 & 51.8 \\
\textbf{AMECON}~\cite{chami2017amecon} & 61.1 & 58.7 & 40.6 & 45.8 & 24.3 & 32.7 & 26.1 & 36.4 & 44.5 \\
\textbf{WhatMakes}~\cite{huh2016makes} & 64.0 & 69.4 & 50.1 & 45.6 & 33.7 & 41.9 & 15.0 &  42.8 & 45.3 \\
\textbf{ISM}~\cite{wu2016ism} & 62.5 & 68.8 & 50.7 & 28.5 & 37.9 & 42.6 & 34.0 & 50.0 & 46.9 \\
%\textbf{GrowingBrain-WA}~\cite{wang2017growing} & 68.4 & 73.1 & 54.7 & 49.3 & 38.4 & 46.5 & 37.5 & 54.8 & 52.8 \\
\textbf{GrowingBrain-RWA}~\cite{wang2017growing} & 69.1 & 74.8 & 55.9 & 50.4 & 40.0 & 48.4 & 38.6 & 56.1 & 54.2 \\
\textbf{FSFT}~\cite{tamaazousti2018universal} & 67.5  & 73.9 & 55.0 & 44.6 & 40.4 & 47.1 & 38.7 & 56.8 & 53.0 \\
%\textbf{MuCaLe (S+G-clu)}~\cite{tamaazousti2017mucale_net} & 69.7 & 76.3 & 54.3 & \textbf{54.7} & 42.4 & 45.0 & 36.9 & 52.8 & 54.0 \\
\textbf{MuCaLe-Net}~\cite{tamaazousti2017mucale_net} & \underline{69.5} & 76.0 & 56.8 & \textbf{54.7} & 41.3 & 48.5 & 35.6 & 54.8 & 54.6 \\
\textbf{MulDiP-Net}~\cite{tamaazousti2018universal} & \textbf{69.8} & 77.5 & \underline{58.3} & 47.9 & \textbf{43.7} & 50.2 & 37.4 & 59.7 & 55.6 \\
\hline
\textbf{FiNet, Random$\dagger$} & 66.4 & 72.4 & 53.2& 51.0 & 39.7 & 46.9 & 35.7 & 55.9 & 52.6 \\
\textbf{FiNet, Cluster$\dagger$} &  66.0 & 73.2 & 54.6 & 50.9 & 40.7 & 47.2  & 36.4 & 55.6 & 53.1 \\
\textbf{FiNet, BUCBAM} & 65.3 & 75.4 & 56.0 & 48.6 & 41.6 & 49.4 & 37.8 & 59.8 & 54.2 \\
\textbf{SpeFiNet, Random$\dagger$} & \textbf{69.8} & 75.7 & 57.5 & \underline{54.6} & 41.2 & 50.0 & 39.8 & 58.3 & 55.9 \\
\textbf{SpeFiNet, Cluster$\dagger$} & 68.6 & \underline{77.9} & 58.1 & 53.9 & 41.3 & \underline{50.5} & \underline{40.8} & \underline{60.1} & \underline{56.4} \\
%\textbf{SpeFiNet, BUCBAM} & 69.2 & 77.0 & 57.4 & \underline{54.6} & \underline{41.6} & 49.9 & \textbf{41.3} & 59.5 & \underline{56.3} \\
\textbf{SpeFiNet, BUCBAM} & 69.1 & \textbf{78.3} & \textbf{59.3} & 54.0 & \underline{42.7} & \textbf{52.0} & \textbf{41.8} & \textbf{61.7} & \textbf{57.4} \\
\hline
\end{tabular}
}
\vspace{-0.3cm}
\caption{
Comparison of our methods (bottom) to the state-of-the-art (top). 
All the methods are trained on the data of ILSVRC* with an AlexNet network and compared in terms of average (Avg) performance on the set of eight target-tasks used in~\cite{tamaazousti2018universal}. 
For each benchmark, we highlight the best score in bold and the second is underlined. 
Methods marked with $\dagger$ are obtained with a parameter cross-validated (on the target-tasks), while our BUCBAM method automatically set this parameter on the source-task.   
Note that MuCaLe-Net, MulDiP-Net and SpeFiNets use representations which dimension is twice other method's.
}
\vspace{-0.3cm}
\label{tab:results_sota}
\egroup
\end{table*}

\begin{table*}[tb!]
\centering
\bgroup
\def\arraystretch{1.2}
\resizebox{\textwidth}{!}{
\begin{tabular}{|l c c c c c c c c c c || c |}
\hline
\multirow{2}{*}{\textbf{Method}} & \textbf{VOC07} & \textbf{CA101} & \textbf{CA256} & \textbf{NWO} & \textbf{MIT67} & \textbf{stACT} & \textbf{CUB} & \textbf{FLO} & \textbf{AIRC} & \textbf{FOOD} & \multirow{2}{*}{\textbf{Avg.}} \\
 & mAP & Acc. & Acc. & mAP & Acc. & Acc. & Acc. & Acc. & Acc. & Acc. & \\
\hline
\hline
\textbf{SpeNet (REF.)}  & 82.7& 91.0 & 78.4 & 70.5 & 64.8 & 72.2 & 59.5 & 80.0 & 49.2 & 47.6 &  69.6 \\
\textbf{GenNet}~\cite{tamaazousti2017mucale_net} & 83.2 & 91.5 & 78.1 & \underline{73.2} & 64.4 & 72.6 & 52.5 & 78.9 & 48.5 & 46.2 & 68.9 \\
\textbf{MulDiP-Net}~\cite{tamaazousti2018universal} & \textbf{84.1} & \textbf{92.7} & \textbf{80.1} & \textbf{73.9} & 66.4 & \underline{74.5} & 61.2 & 82.1 & 53.5 & 49.3 & \underline{71.8} \\
\hline
\textbf{FiNet, Cluster$\dagger$} & 82.5 & 91.8 & 78.8 & 70.0 & 65.8 & 73.2 & 60.9 & 81.9 & 51.9 & 47.8 & 70.5 \\
\textbf{FiNet, Cluster$ (K=16)$} & 81.4 & 91.5 & 77.4 & 69.5 & 64.6 & 72.2 & 58.6 & 81.5 & 52.3 & 47.5 & 69.6 \\
\textbf{FiNet, BUCBAM} & 81.3 & 91.0 & 77.0 & 69.7 & 64.3 & 72.2 & 59.1 & 81.6 & 52.9 & 48.9 & 69.8 \\
%\textbf{FiNet, BUCBAM} & 82.2 & 91.6 & 78.1 & 70.9 & 66.2 & 72.6 & 61.1 & 81.3 & 52.8 & 48.2 & 70.5 \\
%\textbf{FiNet, BUCBAM-SS} & 82.1 & 91.5 & 78.1 & 70.5 & 65.1 & 73.0 & 60.5 & 82.0 & 52.3 & 47.5 & 70.3 \\
\textbf{SpeFiNet, Cluster$\dagger$} & \underline{83.7} & \underline{92.5} & \underline{79.8} & 71.9 & \textbf{66.7} & \textbf{74.8} & \textbf{63.6} & \underline{83.1} & 54.5 & 49.5 & \textbf{72.0} \\ 
\textbf{SpeFiNet, Cluster$ (K=16)$} & 83.3 & 92.2 & 79.6 & 71.9 & \underline{66.6} & 74.1 & 62.5 & 83.0 & \underline{55.1} & \underline{49.8} & \underline{71.8} \\
%\textbf{SpeFiNet, Cluster$ (K=16)$} & 83.3 & \underline{92.2} & 79.6 & 71.9 & \underline{66.6} & \underline{74.1} & \underline{62.5} & \textbf{83.0} & \textbf{55.1} & \underline{49.8} & \underline{71.8} \\
\textbf{SpeFiNet, BUCBAM} & 83.2 & 92.2 & 79.6 & 71.7 & \underline{66.6} & \underline{74.5} & \underline{62.7} & \textbf{83.4} & \textbf{56.1} & \textbf{50.0} & \textbf{72.0} \\
%\textbf{SpeFiNet, BUCBAM} & \underline{83.6} & 92.0 & \underline{80.0} & \underline{72.4} & \textbf{67.5} & \textbf{74.5} & \textbf{63.5} & \underline{82.8} & \underline{54.8} & \textbf{50.3} & \textbf{72.1} \\
%\textbf{SpeFiNet, BUCBAM-SS} & 83.5 & 92.1 & 79.8 & 72.0 & 66.3 & 74.7 & 63.3 & 83.1 & 53.9 & 49.3 & 71.8 \\
\hline
\end{tabular}
}
\vspace{-0.3cm}
\caption{
Comparison of our methods (bottom) to the state-of-the-art (top). 
All the methods are trained on the data of \textit{full} ILSVRC with a DarkNet network and compared in terms of average performance (Avg) on the set of \textit{ten} target-tasks presented in Sec~\ref{sec:settings}. 
For each benchmark, we highlight the best score in bold and the second is underlined. 
Methods marked with $\dagger$ are obtained with a parameter cross-validated on the target-tasks. BUCBAM automatically set this parameter on the source-task.  
}
\vspace{-0.5cm}
\label{tab:results_sota_deep_large}
\egroup
\end{table*}

%--------------------------
% COMPARISON TO SOTA
%--------------------------
\subsection{Comparison to the State-of-the-Art}
\label{sec:comparison_sota}
\vspace{-0.1cm}

We compare the results obtained by our method with those of the literature, in particular, all the methods re-implemented and reported in~\cite{tamaazousti2018universal}. 
For fair comparisons, we followed their training configuration, and trained our method with an AlexNet network and ILSVRC* as source-task. 
Moreover, instead of using our more diverse set of ten target-datasets, we used the eight ones used in their paper. 
The results are reported on Table~\ref{tab:results_sota}. 
We first observe that our methods are always better than the reference method used  in~\cite{tamaazousti2017mucale_net,tamaazousti2018universal,bilen2017universal,rebuffi2017learning,rebuffi2018efficient}, namely SpeNet. 
In particular our best method (BUCBAM) exhibits a boost of 6 points on average, compared to SpeNet. 
Let also note that SpeFiNet is always better than FiNet, itself better than SpeNet, regardless the splitting method. 
More precisely, the BUCBAM splitting method is significantly better than the best Cluster one, without the high cost of cross-validation of the $K$ parameter. 
Compared to state-of-the-art methods, ours achieves the best performances, that is, almost 2 points of improvement compared to the most competitive MulDiP-Net method~\cite{tamaazousti2018universal}, while it surpasses all other methods by more than 3 points. 
A last salient result is the fact that a SpeFiNet (whatever the splitting method) is significantly better than MuCaLe-Net~\cite{tamaazousti2017mucale_net} which has been trained on the best generic categories (manually obtained from categorical-levels~\cite{tamaazousti2016diverse,tamaazousti2017mucale_net}). 
This latter clearly demonstrates than combining features trained on specific categories with those trained on \textit{finer} categories is better than combining them with those trained on \textit{generic} categories. 

Furthermore, since~\cite{tamaazousti2018universal} reported better results with a deeper network (DarkNet) trained on the full ILSVRC, we also implemented our method in the same configuration. 
Since MulDiP-Net provides the best results of the literature on the problem of universality, we only compare to them for this setting. 
The results are reported on Table~\ref{tab:results_sota_deep_large}.  
While the improvement is only slightly better, our method still beats the competitive MulDiP-Net. 
Moreover, a salient observation is that our method tend to be much better than theirs on the fine-grained classification benchmarks, which are more challenging.  
As in the previous setting, SpeFiNet is better than FiNet which is itself better than SpeNet. 
We also compared to the GenNet (which is the generic sub-component of MulDiP-Net~\cite{tamaazousti2018universal}) and we observe that FiNet-BUCBAM is better by 0.9 points. 

\begin{figure*}[tb!]
\begin{center}
    \includegraphics[width=13.0cm]{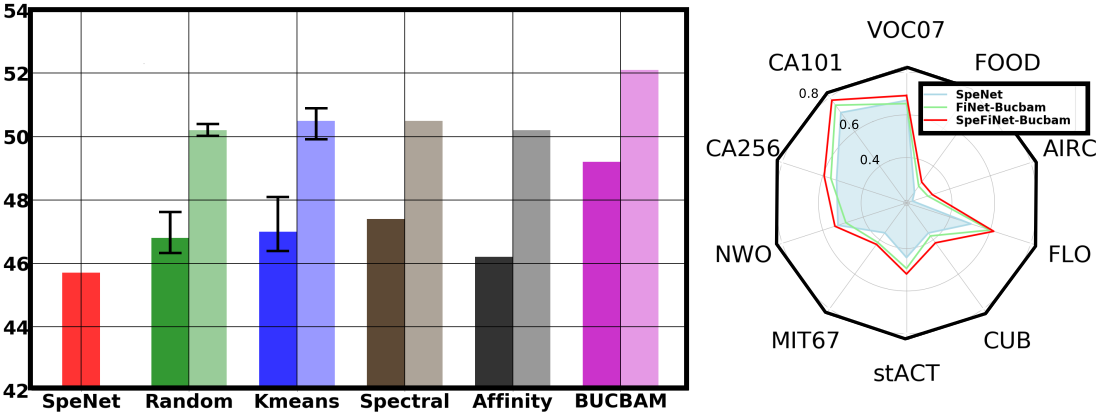}
\end{center}
\vspace{-0.6cm}
\caption{
Comparison of our methods to multiple splitting baselines. 
On the left, we illustrate the average results of each method on the ten target-tasks, were for each splitting method, we illustrate the FiNet in dark color and SpeFiNet in light color. 
On the right, we plot a diagram for the SpeFiNet-BUCBAM, FiNet-BUCBAM and SpeNet methods, their performances on each target-task. 
Best view in color
}
\label{fig:results}
\end{figure*}

%--------------------------
% ANALYSIS AND COMPARISON TO BASELINES
%--------------------------
\subsection{Analysis and Comparison to Baselines}
\label{sec:analysis}

In this section, we perform an in-depth analysis of our method through an ablation study, a comparison to baselines and visualization of some statistics. 
In particular, in supplementary we compared our method to multiple clustering baselines, namely Spectral Clustering and Affinity propagation. 
The former provides a fixed set of clusters per category, while the latter leads to a dynamic set of clusters. 
A summary of results is presented on the left of Fig.~\ref{fig:results}, where we plot, a bar for each method, that represents its average performance on the set of 10 benchmarks described in Sec~\ref{sec:settings}. 
We also tested the Mean-Shift algorithm with many different bandwidth values, but it always led to many clusters with one or two images, and one cluster containing all the remaining images. This setting providing very low results, we did not report them. 
From these results, we observe that our BUCBAM method is better than all the baselines including other existing algorithms. 
Rather than the average performances, in the diagram on the right of Fig.~\ref{fig:results}, we illustrated the \textit{detailed} results (on the ten benchmarks) of the SpeFiNet-BUCBAM, FiNet-BUCBAM and SpeNet methods. 
We clearly observe that the diagram of our SpeFiNet-BUCBAM overlaps FiNet-BUCBAM, which itself overlaps the reference SpeNet method. 
In addition, we provide in supplementary material the detailed results of all the methods on all the target-tasks. 

In Figure~\ref{fig:clusters_vizualisation}, we visualize some of the clusters obtained by each splitting method (random, clustering and BUCBAM). 
To do so, we highlight three clusters for two specific categories (two blocks of three rows of five images). 
On the left, the clusters are determined from a random distribution within the full specific category, leading to clusters that contain its full diversity. 
On the contrary, with the K-means clustering (middle), the clusters exhibits a more coherent aspect. For example, for the \textit{goldfish} category, the $c_3^1$ cluster report close-up views of fish that are rather seen on their profile. We have a similar behaviour for the \textit{banjo} category with cluster $c_1^3$ and $c_2^3$.
With the proposed BUCBAM method (right), the clusters are even more specific than in the K-means case. 
For instance, for the \textit{goldfish} category, we clearly identify a cluster that represents ``many golfishes'' ($c_1^1$), ``on goldfish in a close-up view'' ($c_2^1$) and some images on which the fish tank is visible ($c_3^1$). 
Also for the \textit{banjo} class, we also clearly observe that our method identified a cluster that represents ``person playing banjo'' $c_1^3$ and even ``person playing banjo in a concert'' $c_3^3$. 
Importantly, while the clustering method tend to results in duplicate clusters (\textit{e.g.}, $c^1_2$ with $c^1_3$; $c_1^3$ with $c_2^3$; etc.), ours tend to provide only dissimilar results, thank to our merging process. 

In supplementary material, we also provided some statistics of our method and baselines  (\textit{i.e.}, histograms of average amount of clusters per category, histograms of intra-class variance of the clusters) and more visualizations of the obtained clusters, through the visualization of some images in some clusters and the features of each image of the clusters in a 2D dimensional space, after performing a PCA on their full features. 
This highlights the clear interest of our method, in terms of cluster relevancy and the balance of resulting data, compared to the random and clustering baselines.

\begin{figure}[tb!]
\includegraphics[width=13.0cm]{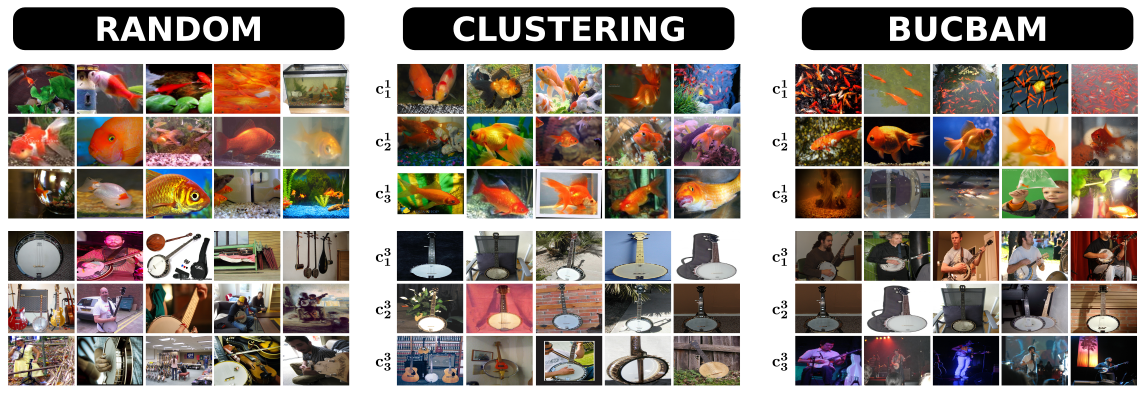}
\vspace{-0.6cm}
\caption{
Illustration of some finer categories obtained from two specific categories (two row blocks) with the different methods: random split (left), Kmeans clustering with K=16 (middle) and our BUCBAM proposal (right). 
In every block, a line (from the three) shows the five most representative images of a cluster at the new finest-level. 
Best view in PDF. 
}
\label{fig:clusters_vizualisation}
\end{figure}

% CONCLUSION
\section{Conclusions}
\vspace{-0.1cm}

In this paper, we tackled the problem of universality of representations with a new method relying on categories that are finer than the most specific ones of the ImageNet hierarchy. 
These last being the finest that are annotated, we proposed a method that automatically add a hierarchical-level to the ImageNet hierarchy. 
A network trained on the categories of such finer-level provides a more universal representation than with the upper levels. In practice, it leads to significantly better results in a transfer-learning scheme, on 10 publicly available datasets from diverse domains.

We also showed that a K-means and, surprisingly, a random partitioning of the leaf nodes of ImageNet already gives interesting results, although below than the proposed approach. 
It nevertheless suggests that the general principle highlighted in this article could be fruitful to design new CNN-based representations that are more universal in a transfer-learning context. 
Furthermore, it should be noted that our principle is neither limited to the ImageNet hierarchy nor to the classification task. 
Indeed, it could be applied to any hierarchy or dataset and on other tasks, such as detection, segmentation or keypoint estimation, as considered in~\cite{wang2018more}.

\appendix
\section*{Supplementary Material}

The following reports some supplementary material that is not required to understand the main article but provide complements or illustrations. 
Hence, the additional elements were produced using the same version of the approach explained in our main paper and include the following items: (i) the detailed characteristics of the datasets used in this paper (Section~\ref{sec:datasets_characteristics}); (ii) detailed results of the comparison of our method with the baselines (Section~\ref{sec:detailed_results}); and finally (iii) some illustrations of the clusters obtained by the different methods as well as some statistics (Section~\ref{sec:clusters_illustration_analysis}). 
%All the figures of this document are best viewed in color.

%--------------------------------
% Datasets: Detailed Characteristics
%--------------------------------
\section{Datasets: Detailed Characteristics}
\label{sec:datasets_characteristics}
\vspace{-0.1cm}

\begin{figure*}[tb!]
\begin{center}
    \includegraphics[width=13.0cm]{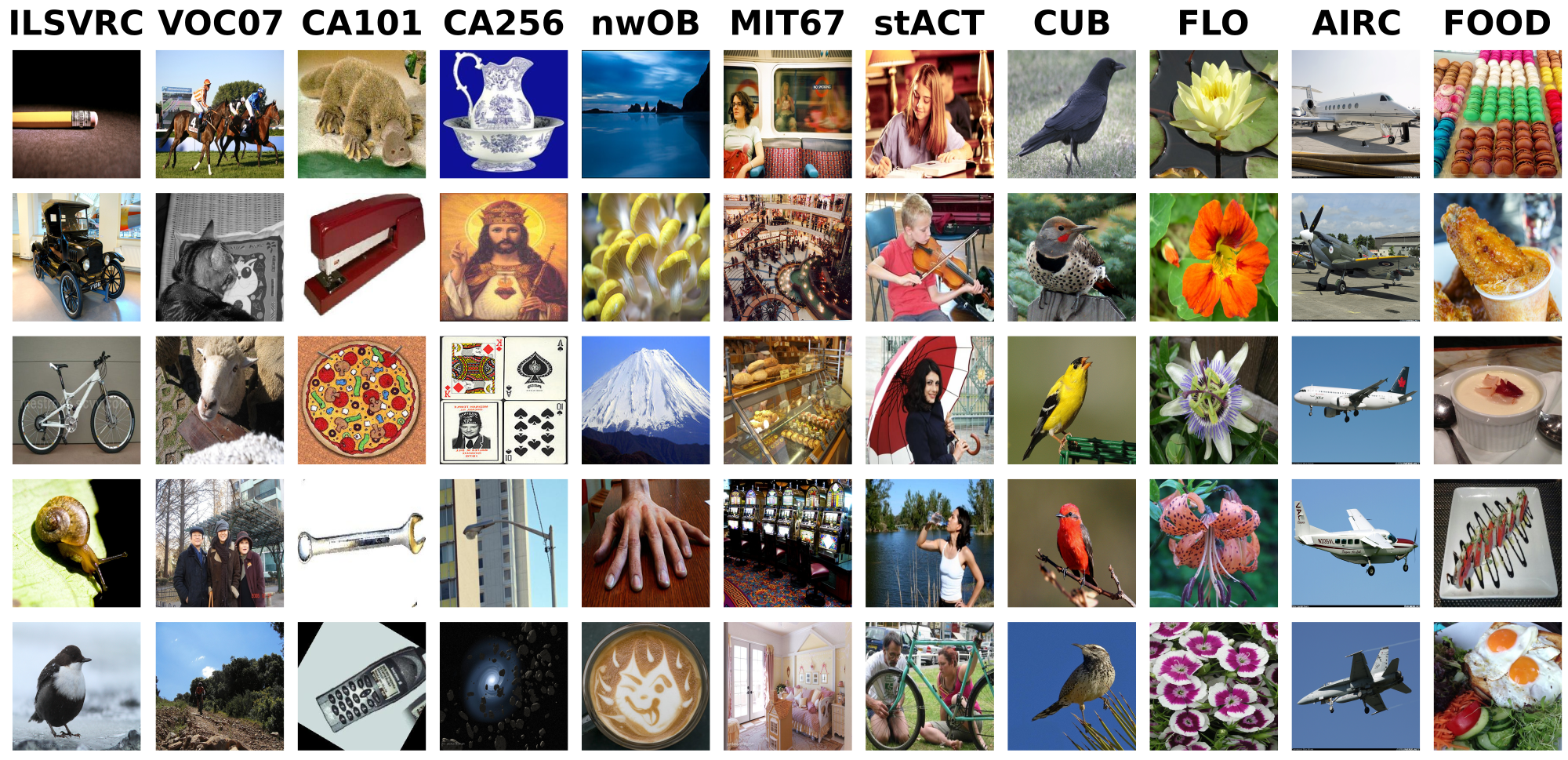}
\end{center}
\vspace{-0.6cm}
\caption{
Illustration of some examples of the ILSVRC source-task as well as the ten target-tasks used to evaluate universality. 
Note the high visual and semantic variability between the different datasets. 
Best view in PDF. 
}
\label{fig:dataset_images_illustration}
\end{figure*}

In Table~\ref{tab:app_datasets}, we report the characteristics of all datasets used in the article to learn CNN on a source-task and to estimate the performances of a universalizing method, that is to say, its performances on a set of target-tasks in the context of transfer-learning. 
For this, we used the most commonly used dataset as source-task, namely ILSVRC~\cite{russakovsky2014imagenet} which is a subset of ImageNet~\cite{deng09imagenet} that contains $1.2$ millions images labeled among 1,000 specific categories. 
We also follow the literature~\cite{tamaazousti2018universal} for fair comparisons and thus used as source-task the ILSVRC* dataset, that corresponds to half of ILSVRC. 
Regarding the target-tasks, we follow the literature~\cite{bilen2017universal,rebuffi2017learning,rebuffi2018efficient,tamaazousti2018universal} and used ten target-datasets in a classification task. 
In particular, here we used benchmarks from various domains, namely objects, actions, scenes, as well as, fine-grained objects like aircrafts, birds, cars and plants. 
In order to show the visual variability of the chosen target-datasets we used to evaluate universalizing methods, we report in Figure~\ref{fig:dataset_images_illustration}, some example images of each of them. 
\vspace{-0.2cm}

\begin{table}[tb!] 
\centering
\bgroup
\def\arraystretch{1.0}
\begin{tabular}{|l c c c c c c c|}
\hline
\textbf{Datasets} & (1) & (2) & (3) & (4) & (5) & (6) & (7) \\ 
\hline
\hline
\textbf{ILSVRC*~\cite{russakovsky2014imagenet}} & objects & 483 & 1,2K & \xmark & 569,000 & 48,299 & Acc.\\
\textbf{ILSVRC~\cite{russakovsky2014imagenet}} & objects  & 1K & 1,2K & \xmark & 1.2M & 50,000 & Acc.\\
\hline
\hline
\hline
\textbf{VOC07}~\cite{everingham2010pascal} & objects & 20 & 250 & \cmark & 5,011 & 4,952 & mAP\\
\textbf{NWO~\cite{nus-wide-civr09}} & objects  & 31 & 700 & \cmark & 21,709 & 14,546 & mAP\\
\textbf{CA101~\cite{fei2006one}} & objects  & 102 & 30 & \xmark & 3,060 & 3,022 & Acc.\\
\textbf{CA256~\cite{griffin2007caltech}} & objects  & 257 & 60 & \xmark & 15,420 & 15,187 & Acc.\\
\textbf{MIT67~\cite{quattoni2009recognizing}} & scenes  & 67 & 80 & \xmark & 5,360 & 1,340 & Acc.\\
\textbf{stACT~\cite{yao2011human}} & actions  & 40 & 100 & \xmark & 4,000 & 5,532 & Acc.\\
\textbf{CUB~\cite{wahCUB_200_2011}} & birds  & 200 & 30 & \xmark & 5,994 & 5,794 & Acc.\\
\textbf{FLO~\cite{nilsback2008automated}} & plants  & 102 & 10 & \xmark & 1,020 & 6,149 & Acc.\\
\textbf{FOOD~\cite{bossard14}} & food  & 101 & 50 & \xmark & 5050 & 5050 & Acc.\\
\textbf{AIRC~\cite{maji13fine-grained}} & airplanes  & 100 & 66 & \xmark & 6,667 & 3,333  & Acc.\\
\hline
\end{tabular}
\vspace{+0.25cm}
\caption{
Detailed descriptive of the different datasets used in the article. 
On top of the table, we describe datasets used as \textit{source-task} and at bottom, those used as \textit{target-task}. 
For each dataset, we detail seven characteristics.  
Each column of the table corresponds to a certain characteristic: (1) domain of the images; (2) amount of categories; (3) average amount of training-images per category; (4) whether the dataset contains multiple categories per image (\cmark) or no (\xmark); (5) amount of training examples; (6) amount of test examples; and (7) the standard evaluation metric (Accuracy and mean Average Precision, respectively denoted by \textbf{Acc.} and \textbf{mAP}). 
Example images of each dataset are presented on Figure~\ref{fig:dataset_images_illustration}. 
}
\label{tab:app_datasets}
\egroup
\end{table}

%--------------------------------
% Comparison to Baselines: Detailed Results
%--------------------------------
\section{Comparison to Baselines: Detailed Results}
\label{sec:detailed_results}
\vspace{-0.1cm}

\begin{figure*}[tb!]
\begin{center}
    \includegraphics[width=7.0cm]{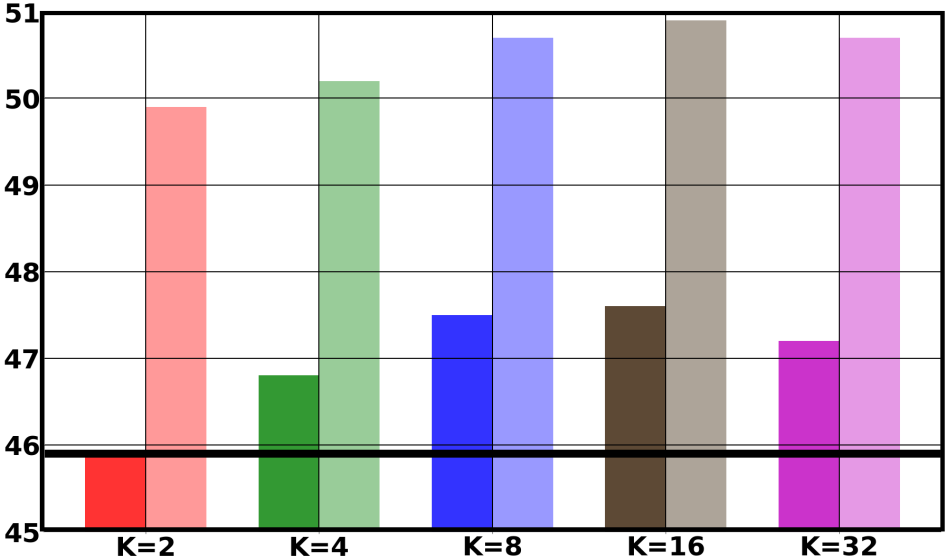}
\end{center}
\vspace{-0.6cm}
\caption{
Average performances of FiNet-Kmeans-$K$ and SpeFiNet-Kmeans-$K$ according different values of $K$. The black line corresponds to the average performance of SpeNet. 
}
\label{fig:cluster_results_according_K}
\end{figure*}

In Figure 3 of the main paper, we reported the synthesis results of the comparison of our methods with several baselines. 
Thus here, we provide detailed results, that is to say, results of all the methods on each benchmark as well as their average performance on all of them. 
This is reported in Table~\ref{tab:results_baselines}. 
Even if already mentioned in the main paper, let recall the most salient results: (i) SpeFiNet is always better than FiNet which is always better than SpeNet, regardless the splitting method; (ii) the proposed BUCBAM splitting method gives better results than the best Kmeans one, at zero cost of parameter cross-validation; and (iii) the proposed BUCBAM is always better than all other methods, especially Random, Spectral and Affinity. 
Additionally, in Figure~\ref{fig:cluster_results_according_K}, we display the average performances of FiNet-Kmeans-$K$ and SpeFiNet-Kmeans-$K$ according different values of $K$, which are compared to the performance of a classical SpeNet.

\begin{table*}[tb!]
\centering
\bgroup
\def\arraystretch{1.2}
\resizebox{\textwidth}{!}{
\begin{tabular}{|l c c c c c c c c c c || c |}
\hline
\multirow{2}{*}{\textbf{Method}} & \textbf{VOC07} & \textbf{CA101} & \textbf{CA256} & \textbf{NWO} & \textbf{MIT67} & \textbf{stACT} & \textbf{CUB} & \textbf{FLO} & \textbf{AIRC} & \textbf{FOOD} & \multirow{2}{*}{\textbf{Avg.}} \\
 & mAP & Acc. & Acc. & mAP & Acc. & Acc. & Acc. & Acc. & Acc. & Acc. & \\
\hline
\hline
%\textbf{Net-G} (Baseline 1) & 65.8 & 70.7 & 50.6 & 51.1 & 35.4 & 43.2 & 17.1 & 43.5 & 47.2 \\
\textbf{SpeNet (REFERENCE)} & 66.8  & 71.1  & 53.2  & 52.5 & 36.0 & 44.3  & 36.1 & 50.5 & 21.6 & 25.0 & 45.7 \\
\textbf{FiNet Random-K2} & 66.6 & 71.8 & 52.7 & 52.6 & 38.9 & 45.9 & 34.4 & 50.9 & 22.7 & 24.0 & 46.0 \\
\textbf{FiNet Random-K4} & 66.8 & 73.1 & 54.4& 52.7 & 38.9 & 47.0 & 34.7 & 53.5 & 23.6 & 25.0 & 47.0 \\
\textbf{FiNet Random-K8} & 67.3 & 71.9 & 54.2& 51.7 & 38.7 & 46.9 & 34.4 & 53.6 & 24.3 & 24.9 & 46.8 \\
\textbf{FiNet Random-K16} & 66.4 & 72.4 & 53.2& 51.0 & 39.7 & 46.9 & 35.7 & 55.9 & 25.2 & 26.2 & 47.3 \\
\textbf{FiNet Kmeans-K2} &  66.3   & 72.2 & 53.7 &  52.1 & 37.7   & 45.4  & 33.8 & 51.5 & 22.6 & 24.2 & 45.9 \\
\textbf{FiNet Kmeans-K4} & 66.9 & 73.0 & 53.0 &  51.4& 39.2   & 46.3  & 35.6 & 54.3 & 22.7 & 25.2 & 46.8 \\
\textbf{FiNet Kmeans-K8} &  66.0 & 73.2 & 54.6 & 50.9 & 40.7 & 47.2  & 36.4 & 55.6 & 24.2 & 26.5 & 47.5 \\
\textbf{FiNet Kmeans-K16} & 64.9 & 73.8 &   54.4 &  50.3    & 39.0   & 47.6  & 36.0 & 56.8 & 26.9 & 26.4 & 47.6 \\
\textbf{FiNet Kmeans-K32} & 63.9 & 72.1 & 53.4 & 48.9  & 40.2   & 47.6  & 36.0 & 57.8 & 26.0 & 26.2 & 47.2 \\
\textbf{FiNet Spectral-K16} & 65.4 & 72.4 & 53.7 & 51.3 & 39.5 & 47.4 & 37.4 & 55.9 & 25.3 & 26.6 & 47.4 \\
\textbf{FiNet Affinity} & 64.5 & 70.5 & 52.1 & 48.5 & 38.7 & 45.9 & 34.8 & 56.0 & 25.3 & 25.5 & 46.2 \\
\textbf{FiNet BUCBAM-AS}   & 66.2 & 72.8 & 54.3 & 51.6 & 39.8 & 46.7 & 35.9 & 56.1 & 25.3 & 25.0 & 47.4 \\
\textbf{FiNet BUCBAM-SS}   & 65.3 & 75.4 & 56.0 & 48.6 & 41.6 & 49.4 & 37.8 & 59.8 & 29.2 & 28.4 & 49.2 \\
\hline
\textbf{SpeFiNet Random-K2} & \textbf{70.0} & 76.0 & 56.8 & \underline{54.9} & 41.3 & 49.4 & 40.4 & 57.4 & 26.4 & 27.9 & 50.0 \\
\textbf{SpeFiNet Random-K4} & 69.4 & 76.0 & 57.9 & \textbf{55.0} & 41.4 & 49.5 & 40.0 & 57.7 & 27.9 & 27.6 & 50.2 \\
\textbf{SpeFiNet Random-K8} & \underline{69.8} & 75.7 & 57.5 & 54.6 & 41.2 & 50.0 & 39.8 & 58.3 & 27.8 & 28.6 & 50.3 \\
\textbf{SpeFiNet Random-K16} & 69.4 & 76.0 & 55.8 & 54.3 & 41.7 & 49.6 & 40.3 & 60.4& 28.0 & 28.9 & 50.4 \\
\textbf{SpeFiNet Kmeans-K2} & 69.2 & 76.3 & 57.1 & 54.7 & 41.2 & 49.0 & 39.0 & 57.8 & 26.4 & 28.3 & 49.9 \\
\textbf{SpeFiNet Kmeans-K4} & 69.6 & 76.2 & 57.1 & 54.2 & 40.1 & 49.6 & 40.8 & 59.1 & 26.7 & 28.3 & 50.2 \\
\textbf{SpeFiNet Kmeans-K8} & 69.1 & 76.6 & 57.8 & 54.1 & 42.2 & 49.6 & 40.4 & 59.8 & 28.3 & 29.3 & 50.7 \\
\textbf{SpeFiNet Kmeans-K16} & 68.6 & \underline{77.9} & \underline{58.1} & 53.9 & 41.3 & \underline{50.5} & 40.8 & 60.1 & \underline{29.8} & 28.4 & \underline{50.9} \\
%\textbf{S+F-(Kmeans, K=16) (300k iter)} & 68.6 & 77.7 & 58.0 & 53.8 &41.2 & 50.4 & 40.9 & 60.1 & 56.3 \\
\textbf{SpeFiNet Kmeans-K32} & 68.4 & 76.6 &  57.5 & 53.5  & \underline{42.3}   & 50.3  & 40.4 & \underline{60.7} & 28.8 & 29.0 & 50.7 \\
\textbf{SpeFiNet Spectral-K16} & 68.9 & 76.0 & 57.0 & 54.4 & 41.0 & 49.5 & \underline{41.2} & 59.3 & 28.9 & \underline{29.4} & 50.5 \\
\textbf{SpeFiNet Affinity} & 68.6 & 75.7 & 57.1 & 53.4 & 41.9 & 49.1 & 40.1 & 59.5 & 27.6 & 28.6 & 50.2 \\
\textbf{SpeFiNet BUCBAM-AS}   & 69.4 & 76.3 & 57.6 & 54.4 & 41.1 & 49.6 & 40.2 & 60.1 & 28.4 & 28.7 & 50.6 \\
\textbf{SpeFiNet BUCBAM-SS}   & 69.1 & \textbf{78.3} & \textbf{59.3} & 54.0 & \textbf{42.7} & \textbf{52.0} & \textbf{41.8} & \textbf{61.7} & \textbf{31.4} & \textbf{30.8} & \textbf{52.1} \\
\hline
\end{tabular}
}
\vspace{-0.3cm}
\caption{
Comparison of the proposed universalizing methods (BUCBAM) to baselines (Random, Kmeans, Spectral and Affinity) and the reference one (SpeNet). 
The comparison is carried in a transfer-learning scheme on the ten target-datasets presented in Section~\ref{sec:datasets_characteristics}, for which we report the performances of the methods on each dataset (with standard evaluation-metrics) and the average performance on all the benchmarks (in the last column). 
All the methods have been learned with the same architecture (AlexNet) on the same initial source-problem (ILSVRC*). 
As in all the Tables of the main paper, for each dataset, we highlight the score of the best method in bold and those of the second is underlined. 
}
\label{tab:results_baselines}
\egroup
\end{table*}

\begin{figure*}[tb!]
\begin{center}
    \includegraphics[width=13.0cm]{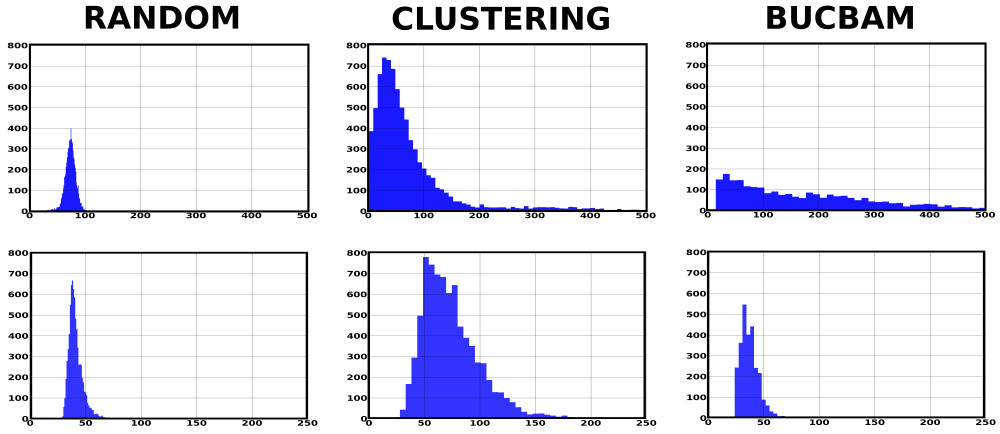}
\end{center}
\vspace{-0.6cm}
\caption{
Illustration of some statistics of the different splitting methods, namely random, clustering and BUCBAM. 
On top, we illustrate the \textit{histogram of the amount of samples per cluster} for all the specific categories of ILSVRC, with the different splitting methods. 
At bottom, we illustrate the \textit{histogram of the intra-class variance of the cluster}. On vertical axis is the number of cluster.
}
\label{fig:histograms_amount_images_per_cluster}
\end{figure*}

\begin{figure}[tb!]
\includegraphics[width=13.0cm]{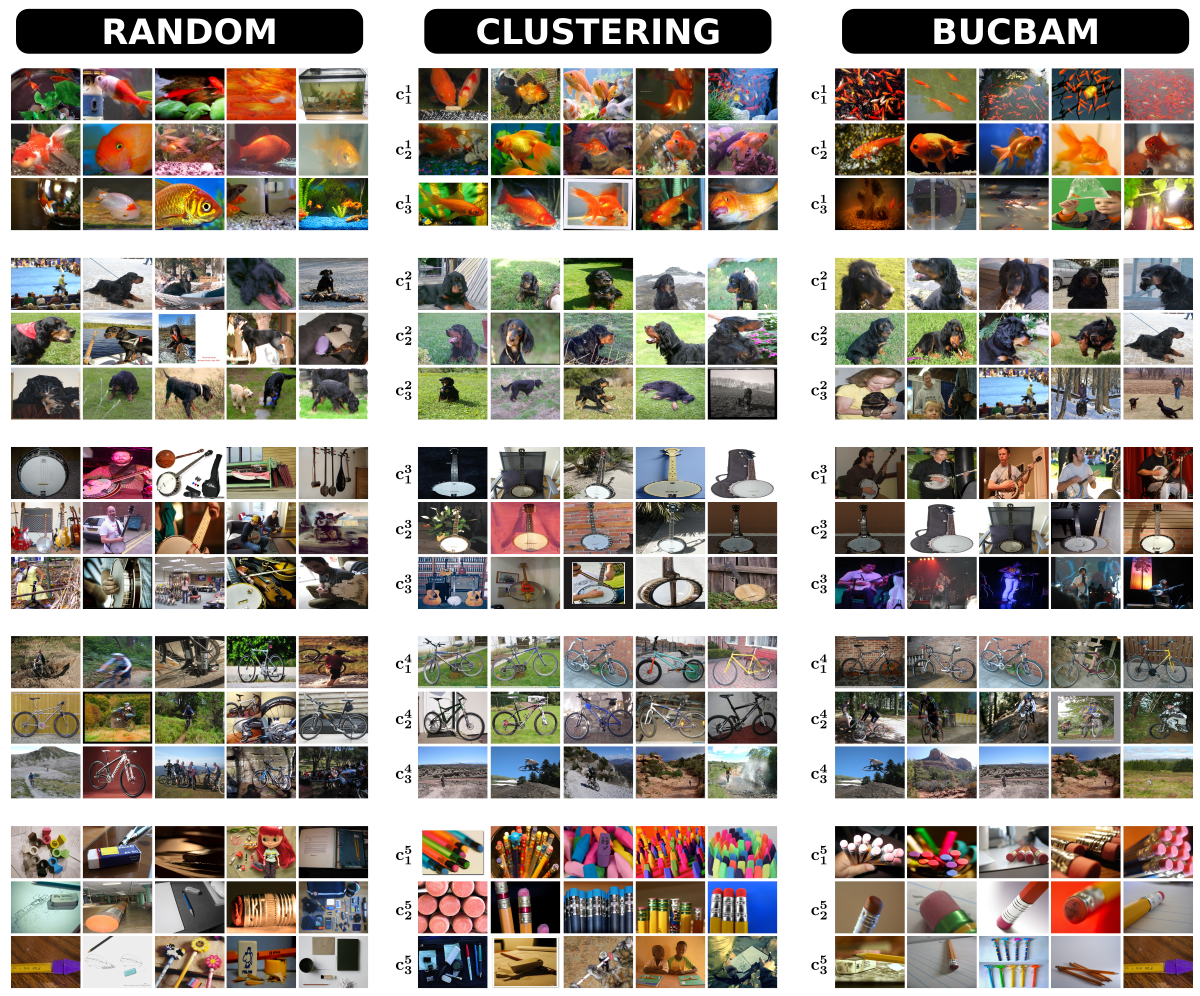}
\vspace{-0.6cm}
\caption{
Illustration of the finer categories obtained from five specific categories (five row blocks) with the different methods: random split (left), Kmeans clustering with K=16 (middle) and our BUCBAM proposal (right). 
In each of the five block, each line shows the five most representative images of a cluster at the new finest-level. 
Best view in PDF. 
}
\label{fig:clusters_vizualisation}
\end{figure}

\begin{figure}[tb!]
\includegraphics[width=13.0cm]{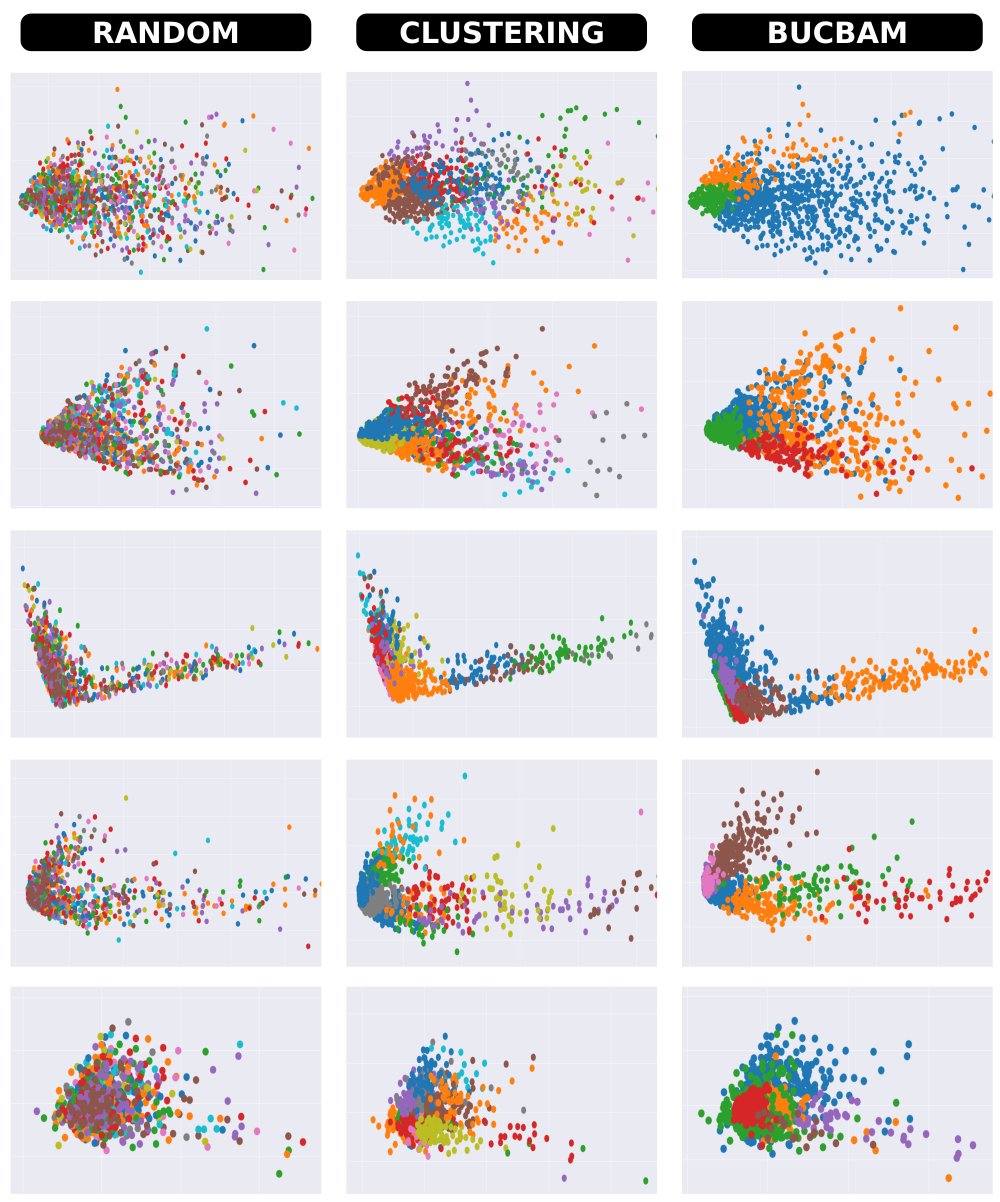}
\vspace{-0.6cm}
\caption{
Illustration of the projection of the samples of the clusters on their two first principal components (obtain by PCA) for the three different methods: random split (left) clustering with Kmeans, K=16 (middle) and our BUCBAM proposal (right). 
In each graph, each color represents a certain cluster. 
For all the methods, each line represents the instances of a specific category, that is the same than the one represented at the same line in Figure~\ref{fig:clusters_vizualisation}. 
Note that, in contrast to other methods, ours provide clusters with \textit{different} amount of colors for each specific category. Best view in PDF.
}
\label{fig:pca_clusters_vizualisation}
\end{figure}

%--------------------------------
% Splitting Methods: Illustration of Clusters and Statistics
%--------------------------------
\section{Splitting Methods: Statistics and Visualization}
\label{sec:clusters_illustration_analysis}

In this section we illustrate some interesting properties of the random, cluster and BUCBAM splitting methods. 
In particular, we first highlight some statistics in Figure~\ref{fig:histograms_amount_images_per_cluster}. 
Indeed, on top we plot, for each method, the \textit{histogram of amount of images per cluster} for all the specific categories of the initial dataset.  
Note that, the more the histogram forms a pointed spike, the more the data are balanced. 
Here we clearly observe that the random splitting method provides the most balanced data, while in contrast other methods tend to contain clusters of various sizes.  highly imbalanced data, 
However, if they provide imbalanced data, it is important to mention that clustering and BUCBAM provide \textit{relevant} clusters that are based on the semantic encoded on the image features, which contain thus samples that are more visually similar. 
Let also note that, compared to the histogram of clustering that has a long tail and starts at near-zero, the histogram of BUCBAM is more flat and starts around 20, meaning that no tail is modeled in the data and very small clusters are not considered. 

At bottom of Figure~\ref{fig:histograms_amount_images_per_cluster}, we reported the \textit{histogram of intra-class variance of the clusters} obtained from all the specific categories, by the three splitting methods. 
In this, it is important to note that a small width of the histogram means that the set of  clusters contains very similar samples. 
While random provides the smallest width of the peak, it is necessary to observe this is due to the fact that it has almost the same amount of images per cluster, thus it is not relevant. In contrast, BUCBAM that provides a large set of amount of categories, also provides a width of the peak that is lowest than clustering, meaning that it provides clusters with more similar samples.  

While previously, we reported some global statistics of the resulting clusters from the different splitting methods, here we rather show the most representative samples of the clusters obtained by each splitting method. 
Indeed, this is reported on Figure~\ref{fig:clusters_vizualisation}, on which we highlight three clusters (three rows of images) for five specific categories (five blocks of three rows of images). 
On the left, the clusters are determined from a random distribution within the full specific category, leading to clusters that contain its full diversity. 
Note by the way, how diverse the specific categories are, and imagine how generic categories (used in~\cite{tamaazousti2016diverse,tamaazousti2017mucale_net}) could be, which may explain   why the GenNet of~\cite{tamaazousti2017mucale_net} may not provide good results (since it is hard from it to discover relevant features). 
On the contrary, with our splitting method and more precisely the K-means clustering (middle), the clusters exhibits a more coherent aspect. For example, for the \textit{goldfish} category, the $c_3^1$ cluster report close-up views of fish that are rather seen on their profile. We have a similar behaviour for the \textit{bicycle} category with cluster $c_2^4$ and $c_3^4$.
With the method we propose (right), the clusters are even more specific than in the K-means case. For instance, for the \textit{goldfish} category, we clearly identify a cluster that represents ``many golfishes'' ($c_1^1$), ``on goldfish in a close-up view'' ($c_2^1$) and some images on which the fish tank is visible ($c_3^1$). 
Also for the \textit{banjo} class, we also clearly observe that our method identified a cluster that represents ``person playing banjo'' $c_1^3$ and even ``person playing banjo in a concert'' $c_3^3$. 
Importantly, while the clustering method tend to results in duplicate clusters (\textit{e.g.}, $c^1_2$ with $c^1_3$; $c_1^3$ with $c_2^3$; $c_1^4$ with $c_2^4$ etc.), ours tend to provide only dissimilar results, thank to our merging process. 

Finally, we also computed a principal component analysis of the representations of each specific category and projected the vectors on the first two principal components, keeping a different color for each (new) finer category (Figure~\ref{fig:pca_clusters_vizualisation}). As expected, with the random split, the vectors are uniformly distributed while the two other methods tend to form some groups. Although these results are qualitative, one can see that the proposed BUCBAM method exhibits slightly more grouped points than the K-means.

\end{document}